\documentclass[oneside,10pt]{article} 

\usepackage{amssymb}
\usepackage{natbib}
\usepackage{graphicx}
\usepackage{float}
\usepackage{array}
\usepackage[cmex10]{amsmath}

\DeclareMathAlphabet{\mathpzc}{OT1}{pzc}%
                                 {m}{it}

\newcommand{\mybox}[1] {\begin{center} \fbox{\parbox{0.80\linewidth}{#1}} \end{center}}

\newtheorem{theorem}{Theorem}[section]

\newtheorem{proposition}[theorem]{Proposition}

\newenvironment{proof}[1][Proof]{\begin{trivlist}
\item[\hskip \labelsep {\bfseries #1}]}{\end{trivlist}}

\newcommand{\qed}{\nobreak \ifvmode \relax \else
      \ifdim\lastskip<1.5em \hskip-\lastskip
      \hskip1.5em plus0em minus0.5em \fi \nobreak
      \vrule height0.75em width0.5em depth0.25em\fi}

\begin{document}

\title{Efficient Marginal Likelihood Computation for Gaussian Process Regression}

\title{Efficient Marginal Likelihood Computation for Gaussian Processes and Kernel Ridge Regression}
\author{Andrea Schirru$^{1}$, Simone Pampuri$^1$, Giuseppe De Nicolao$^1$ and Se\a'an McLoone$^2$\\
$^{(1)}$ University of Pavia, Italy, $^{(2)}$ National University of Ireland, Maynooth\\
\texttt{andrea.schirru@unipv.it}, \texttt{simone.pampuri@unipv.it},\\
\texttt{giuseppe.denicolao@unipv.it}, \texttt{ sean.mcloone@eeng.nuim.ie} }

\maketitle

\begin{abstract} Statistical learning methodologies are nowadays applied in several scientific and industrial settings; in general, any application that strongly relies on data and would benefit from predictive capabilities is a candidate for such approaches. In a Bayesian learning setting, the posterior distribution of a predictive model arises from a trade-off between its prior distribution and the conditional likelihood of observed data. Such distribution functions usually rely on additional hyperparameters which need to be tuned in order to achieve optimum predictive performance; this operation can be efficiently performed in an Empirical Bayes fashion by maximizing the posterior marginal likelihood of the observed data. Since the score function of this optimization problem is in general characterized by the presence of local optima, it is necessary to resort to global optimization strategies, which require a large number of function evaluations. Given that the evaluation is usually computationally intensive and badly scaled with respect to the dataset size, the maximum number of observations that can be treated simultaneously is quite limited. In this paper, we consider the case of hyperparameter tuning in Gaussian process regression. A straightforward implementation of the posterior log-likelihood for this model requires $\mathcal{O}(N^3)$ operations for every iteration of the optimization procedure, where $N$ is the number of examples in the input dataset. We derive a novel set of identities that allow, after an initial overhead of $\mathcal{O}(N^3)$, the evaluation of the score function, as well as the Jacobian and Hessian matrices, in $\mathcal{O}(N)$ operations. We prove how the proposed identities, that follow from the eigendecomposition of the kernel matrix, yield a reduction of several orders of magnitude in the computation time for the hyperparameter optimization problem. Notably, our solution provides computational advantages even with respect to state of the art approximations that rely on sparse kernel matrices. A simulation study is used to validate the presented results. In conclusion, it is shown how the contribution of the present paper is the new state of the art for exact hyperparameter tuning in Gaussian process regression.
\end{abstract}

\newpage

\section*{INTRODUCTION}

Statistical learning methodologies are nowadays applied in a wide spectrum of practical situations: notable examples can be easily found in economics \citep{demiguel2009generalized}, advanced manufacturing \citep{lynn2009virtual}, biomedical sciences \citep{dreiseitl2001comparison}, robotics \citep{vijayakumar2002statistical}, and generally any data-intensive application that would benefit from reliable prediction capabilities. In short, the goal of statistical learning is to characterize the behavior of a phenomenon from a representative \emph{training set} of observations. Such characterization, referred to as a \emph{model}, then allows a probabilistic characterization (or prediction) of unobserved data \citep{hastie2005elements}. In the following, let

$$
\mathcal{S} = \{\mathbf{X} \in \mathbb{R}^{N \times P}, \mathbf{y} \in \mathbb{R}^{N}\}
$$

\noindent
denote a training set consisting of an \emph{input matrix}, $\mathbf{X}$, and a target \emph{output vector}, $\mathbf{y}$. Typically, the former is associated with easily obtainable information (for instance, sensor readings from an industrial process operation), while the latter relates to more valuable data (such as a qualitative indicator of the aforementioned process, resulting from a costly procedure). The goal is then to find a probabilistic model $f(\cdot)$ such that, given a new observation $\mathbf{x}_{new} \notin \mathcal{S}$, $f(\mathbf{x}_{new})$ will yield an accurate probabilistic description of the unobserved $y_{new}$.  

Here, we focus on the well known Bayesian learning paradigm, according to which the structure of the model $f$ arises as a trade-off between an $\emph{a priori}$ distribution $p(f)$, reflecting the prior prejudice about the model structure, and a $\emph{likelihood}$ distribution $p(\mathbf{y}|f)$, that can be seen as a measure of the approximation capabilities of the model \citep{Bernardo94}. In more rigorous terms, Bayes' theorem states that

$$
p(f|\mathbf{y}) \propto p(\mathbf{y}|f)p(f)
$$
\noindent
where $p(f|\mathbf{y})$ is referred to as \emph{a posteriori} distribution. In general the prior and likelihood distributions rely on a set of unknown \emph{hyperparameters} $\theta$ that must be estimated as part of the training process; it is well known that an optimal tuning of $\theta$ is critical to achieving maximum prediction quality. It is common to employ non-informative hyperprior distributions for $\theta$ (i.e., $p(\theta) = \mathrm{const}$ for every feasible $\theta$) and then to maximize with respect to $\theta$ the \emph{residual likelihood distribution} 

\begin{equation}
p(\mathbf{y}|\theta) \label{evidence}
\end{equation}

\noindent
obtained by marginalizing $p(\mathbf{y}|f, \theta)$ with respect to $f$. There are two major challenges to achieving this result: \textbf{(i)} score function evaluation is usually computationally demanding and badly scaled with respect to the size of $\mathcal{S}$; and \textbf{(ii)} in nontrivial cases, the score function has multiple local maxima \citep{carlin1997bayes}. While iterative global optimization techniques can be employed to overcome these challenges, in general, they require a large number of function evaluations/iterations to converge \citep{karaboga2008performance}. It is immediately apparent that the combination of these issues can lead to computationally intractable problems for many real life datasets. Remarkably, this problem is so relevant that a number of approximate formulations have been researched in the last decade in order to reduce the computational demand: see for example \citep{smola2001sparse} and \citep{lázaro2010sparse}.

In this paper, a novel set of equations is developed for the efficient computation of the marginalized log-distribution, as well as its Jacobian and Hessian matrices, for the specific case of Gaussian process regression. The proposed computation, which exploits an eigendecomposition of the Kernel matrix, has an initial overhead of $\mathcal{O}(N^3)$, after which it is possible to calculate the quantities needed for the iterative maximization procedure with a complexity of $\mathcal{O}(N)$ per iteration. Notably, this result makes the proposed solution amenable even in comparison to the aforementioned approximations.

The remainder of the paper is organized as follows: The problem of tuning hyperparameters via marginal likelihood maximization in the cases of Gaussian process regression is introduced in Section \ref{sec1}. Section \ref{sec2} then derives and presents the main results of the paper, and states the computational advantage with respect to the state of the art. The results are validated with the aid of a simulation study in Section \ref{sec3}. Finally, following the conclusions of the paper, Appendix A provides mathematical proofs of the main results.

\section{PROBLEM STATEMENT} \label{sec1}

In this section, the hyperparameter optimization problem for Gaussian process regression is presented. In the following, let

$$
G(\mathbf{\mu}, \mathbf{\Sigma}; \mathbf{x}) = \frac{1}{(2\pi)^{n/2}|\mathbf{\Sigma}|^{1/2}}\displaystyle{\mathrm{exp}\left({-\displaystyle\frac{1}{2} (\mathbf{x} - \mathbf{\mu})\mathbf{\Sigma}^{-1}(\mathbf{x} - \mathbf{\mu})'}\right)}
$$

\noindent be the probability distribution function of a multivariate Gaussian random variable with expected value $\mathbf{\mu}$ and covariance matrix $\mathbf{\Sigma}$, and let 

\begin{equation}
\mathcal{L} = -\frac{1}{2}\log{|\mathbf{\Sigma}|} - \frac{1}{2}(\mathbf{x} - \mathbf{\mu})\mathbf{\Sigma}^{-1}(\mathbf{x} - \mathbf{\mu})' \label{logl}
\end{equation}

\noindent be the log-likelihood of such a distribution up to an additive constant. Furthermore, consider the following theorem that was first defined in \citep{miller1964miller}.

\begin{theorem} \label{theorem1} Let $\mathbf{A} \in \mathbb{R}^{s\times s}$, $\mathbf{a} \in \mathbb{R}^s$, $\mathbf{B} \in \mathbb{R}^{t \times t}$, $\mathbf{b} \in \mathbb{R}^t$ and $\mathbf{Q} \in \mathbb{R}^{s \times t}$. Let $\mathbf{x} \in \mathbb{R}^t$ be an input variable. It holds that

\begin{equation*}
G(\mathbf{a}, \mathbf{A}; \mathbf{Q}\mathbf{x})G(\mathbf{b}, \mathbf{B}; \mathbf{x}) = G(\mathbf{a}, \mathbf{A} + \mathbf{Q}\mathbf{B}\mathbf{Q}'; \mathbf{b})G(\mathbf{d}, \mathbf{D}; \mathbf{x})
\end{equation*}

\noindent with
\begin{eqnarray*}
\mathbf{D} &=& (\mathbf{Q}'\mathbf{A}^{-1}\mathbf{Q} + \mathbf{B}^{-1})^{-1}\\
\mathbf{d} &=& \mathbf{b} + \mathbf{D}\mathbf{Q}'\mathbf{A}^{-1}(\mathbf{a} - \mathbf{Q}\mathbf{b})
\end{eqnarray*}

$\qed$

\end{theorem}

\noindent Given a training set $\mathcal{S}$, let $\mathbf{K} \in \mathbb{R}^{N \times N}$ be the full kernel matrix whose $[i, j]$ entry is defined as

\begin{equation}
\mathbf{K}[i, j] = \mathcal{K}(\mathbf{x}_i, \mathbf{x}_j) \label{kernelmatrix}
\end{equation}

\noindent where $\mathcal{K(\cdot,\cdot)}$ is a suitable positive definite kernel function and $\mathbf{x}_k$ is the $k$-th row of $\mathbf{X}$. Furthermore, consider a parametrization of the model $f$ such that

\begin{equation}
f(\tilde{x};\ \mathbf{c}) = \mathbf{k_{\tilde{x}}}\mathbf{c} + \epsilon \label{modelstruct}
\end{equation}

\noindent where $\epsilon \sim N(0, \sigma^2)$ is a noise term, $\mathbf{k_{\tilde{x}}} \in \mathbb{R}^{1 \times N}$ is a row vector whose $j$-th element is

$$
\mathbf{k_{\tilde{x}}}[j] = \mathcal{K}(\mathbf{\tilde{x}}, \mathbf{x}_j)
$$

\noindent and $\mathbf{c}$ is the unknown parameter vector. This model structure relies on RKHS (Reproducing Kernel Hilbert Spaces) theory, the details of which are beyond the scope of this paper; the interested reader is referred to \citep{schölkopf2002learning} and \citep{shawe2004kernel}. In the following, we restrict consideration to the case where the observation likelihood is derived from equation (\ref{modelstruct}) as

\begin{equation}
\mathbf{y}|\mathbf{c} \sim N(\mathbf{K}\mathbf{c}, \sigma^2\mathbf{I}) \label{likelihood}
\end{equation}

\noindent and the prior distribution of $\mathbf{c}$ is

\begin{equation}
\mathbf{c} \sim N(0, \lambda^2\mathbf{K}^{-1}) \label{prior}
\end{equation}

The hyperparameter $\lambda^2$ accounts for the variability of the coefficient vector $\mathbf{c}$, while $\sigma^2$ is  the output variance defined in equation (\ref{modelstruct}). Notably, equations (\ref{likelihood}) and (\ref{prior}) describe Gaussian process regression \citep{rasmussen2004gaussian} as well as the Bayesian interpretation of kernel ridge regression. Applying Bayes' theorem yields

$$
p(\mathbf{c}|\mathbf{y}) \propto p(\mathbf{y}|\mathbf{c})p(\mathbf{c})
$$

\noindent and the posterior distribution of $\mathbf{c}$ can be readily computed using Theorem \ref{theorem1} as

\begin{eqnarray}
\mathbf{c}|\mathbf{y} &\sim & N(\mathbf{\mu_c}, \mathbf{\Sigma_c}) \label{postc}\\
\mathbf{\mu_c} &=& \left(\mathbf{K} + \frac{\sigma^2}{\lambda^2}\mathbf{I}\right)^{-1}\mathbf{y} \label{mu_c}\\
\mathbf{\Sigma_c} &=& \sigma^2 \left(\mathbf{K} + \frac{\sigma^2}{\lambda^2}\mathbf{I}\right)^{-1}\mathbf{K}^{-1} \label{Sigma_c} \end{eqnarray}

\noindent By combining (\ref{modelstruct}) and (\ref{postc}), the prediction distribution for $\tilde{y}$ is

$$
f(\mathbf{\tilde{x}};\ \mathbf{c}|\mathbf{y}) \sim N(\mathbf{k_{\tilde{x}}}\mathbf{\mu_c}, \mathbf{k_{\tilde{x}}}\mathbf{\Sigma_c}\mathbf{k_{\tilde{x}}}' + \sigma^2)
$$

\noindent The posterior distribution of $\mathbf{y}$, obtained by marginalizing with respect to $\mathbf{c}$,  reads then

$$
\mathbf{y}|(\sigma^2, \lambda^2) \sim N\left(\mathbf{\mu_y}, \mathbf{\Sigma_y}\right) 
$$

\noindent with
\begin{eqnarray}
\mu_\mathbf{y} &=& \mathbf{K}\mu_c = \mathbf{K}\left(\mathbf{K} + \frac{\sigma^2}{\lambda^2}\mathbf{I}\right)^{-1}\mathbf{y} \label{muY}\\
\mathbf{\Sigma_y} &=& \mathbf{K}\mathbf{\Sigma_c}\mathbf{K} + \sigma^2\mathbf{I} = \sigma^2\left(\mathbf{K}\left(\mathbf{K} + \frac{\sigma^2}{\lambda^2}\mathbf{I}\right)^{-1} + \mathbf{I}\right) \label{SigmaY}
\end{eqnarray}

It is apparent that, after the marginalization with respect to $\mathbf{c}$, the posterior distribution $p(\mathbf{y}|(\sigma^2, \lambda^2) )$ depends only on the unknown hyperparameters $\sigma^2$ and $\lambda^2$. The optimal values for these hyperparameters arise as the solution to the optimization problem

\begin{equation}
\{\hat\sigma^2, \hat\lambda^2\} = \arg \underset{\sigma^2, \lambda^2}{\max}\ p(\mathbf{y}|(\sigma^2, \lambda^2)) \label{scorefun0}
\end{equation}

\noindent subject to the constraints
\begin{equation}
\begin{cases}
\sigma^2 > 0 \\
\lambda^2 > 0 \label{const}
\end{cases}
\end{equation}

\noindent Equation (\ref{scorefun0}) is usually log-transformed for numerical reasons and converted to a minimization problem, that is:

\begin{equation}
\{\hat\sigma^2, \hat\lambda^2\} = \arg \underset{\sigma^2, \lambda^2}{\min}\ \mathcal{L}_y \label{scorefun1}
\end{equation}

\noindent again subject to (\ref{const}), where
$$
\mathcal{L}_y := \mathcal{L}_y(\sigma^2, \lambda^2) = -2\log p(\mathbf{y}|(\sigma^2, \lambda^2))
$$

\noindent Notably, equation (\ref{scorefun1}) defines a global optimization problem whose score function derives from equation (\ref{logl}) as

\begin{equation}
\mathcal{L}_y = -2\log(p(\mathbf{y}|(\sigma^2, \lambda^2)) = \log{|\mathbf{\Sigma_y}|} + (\mathbf{\mu_y} -\mathbf{y})\mathbf{\Sigma_y}^{-1}(\mathbf{\mu_y} - \mathbf{y})' \label{loglfull}
\end{equation}

\subsection{Optimization Strategies for Likelihood Maximization} \label{oldoptim}

In order to minimize the nonconvex score function $\mathcal{L}_y$, it is necessary to resort to a global optimization procedure. The optimal solution is typically obtained in two steps: first, an approximate global minimizer is found by means of a global optimization procedure; notable examples include grid search strategies, Particle Swarm Optimization (PSO) and Genetic Algorithms \citep{petelin2011optimization}. Usually, such methods rely only on evaluations of the score function itself, and not on its derivatives, and can generally avoid being trapped in local minima. The approximate minimizer obtained is subsequently used as the starting point for a descent algorithm (such as the Steepest Descent or Newton-Raphson method) that exploits the Jacobian (and possibly the Hessian) of the score function to converge to a local optimal solution. To summarize,

\begin{itemize}
\item The global optimization step requires a large number of evaluations of the score function $\mathcal{L}_y$ in order to span the parameter space in a dense way
\item The local optimization step requires the evaluation of $\mathcal{L}_y$, and its Jacobian (and Hessian, if needed by the method) once per iteration, but usually only needs a small number of iterations to converge
\end{itemize}

\noindent In order to characterize the computational costs associated with each iteration, consider the identity

$$
(\mathbf{\mu_y} - \mathbf{y}) = \frac{1}{\sigma^2}(\mathbf{\Sigma_y} - 2\sigma^2\mathbf{I})\mathbf{y}
$$

\noindent This allows $\mathcal{L}_y$ to be conveniently expressed as

\begin{equation}
\mathcal{L}_y = \log{|\mathbf{\Sigma_y}|} + \frac{1}{\sigma^4}\mathbf{y}'\mathbf{\Sigma_y}\mathbf{y} + 4\mathbf{y}'\mathbf{\Sigma_y}^{-1}\mathbf{y} - 4\frac{\mathbf{y}'\mathbf{y}}{\sigma^2}
\end{equation}

\noindent It is apparent, given equation (\ref{SigmaY}), that it is necessary to compute the inverse of an $N \times N$ matrix in order to evaluate $\mathcal{L}_y$, an operation that has a computational complexity of $\mathcal{O}(N^3)$. The Jacobian computation also has an $\mathcal{O}(N^3)$ bottleneck due to the term

$$
\frac{\partial{\log|\mathbf{\Sigma_y}|}}{\partial \sigma^2} = \operatorname{tr}\left(\mathbf{\Sigma_y}^{-1}\frac{\partial{\mathbf{\Sigma_y}}}{\partial \sigma^2}\right)
$$

At this point, having both $\mathbf{\Sigma_y}$ and $\mathbf{\Sigma_y}^{-1}$ stored in memory, it is possible to compute the Hessian matrix with a complexity of $\mathcal{O}(N^2)$. In conclusion, a computational complexity of $\mathcal{O}(N^3)$, along with a memory storage of $\mathcal{O}(N^2)$, is required to perform each iteration of both the global and local optimization steps. This poses a severe constraint on the maximum size of dataset that can be employed to solve the hyperparameter optimization problem: for this reason, state of the art techniques for overcoming such complexity constraints commonly rely on sparse approximations \citep{quiñonero2005unifying}.

\section{MARGINAL LOGLIKELIHOOD VIA KERNEL DECOMPOSITION} \label{sec2}

Section \ref{sec1} showed that the marginalized likelihood maximization problem for Gaussian process regression is characterized by local optimality and high computational complexity. In practical applications this represents a strict constraint on the size of the training dataset $\mathcal{S}$. In this section, a novel set of identities to efficiently compute the quantities involved in the optimization problem are derived and discussed. Specifically, in order to perform both the global and local optimization steps described in the previous section, the following quantities need to be repeatedly computed for different parameter values:

\begin{itemize}
\item Score function $\mathcal{L}_y(\sigma^2, \lambda^2)$
\item First order derivatives $\displaystyle\frac{\partial\mathcal{L}_y(\sigma^2, \lambda^2)}{\partial\sigma^2}$ and $\displaystyle\frac{\partial\mathcal{L}_y(\sigma^2, \lambda^2)}{\partial\lambda^2}$
\item Second order derivatives $\displaystyle\frac{\partial^2\mathcal{L}_y(\sigma^2, \lambda^2)}{\partial^2\sigma^2}$, $\displaystyle\frac{\partial^2\mathcal{L}_y(\sigma^2, \lambda^2)}{\partial^2\lambda^2}$ and $\displaystyle\frac{\partial^2\mathcal{L}_y(\sigma^2, \lambda^2)}{\partial\sigma^2\partial\lambda^2}$
\end{itemize}

\noindent In the following, let $\mathbf{U} \in \mathbb{R}^{N \times N}$ be the eigenvector matrix and $\mathbf{S}  \in \mathbb{R}^{N \times N}$ be the diagonal eigenvalues matrix such that

\begin{equation}
\mathbf{USU'} = \mathbf{K}
\end{equation}

\noindent where $\mathbf{K}$ is defined in equation (\ref{kernelmatrix}) and let $s_i$ be the $i$-th ordered eigenvalue of $\mathbf{K}$. Furthermore, let $\tilde{y}_i$ be the $i$-th element of the vector

\begin{equation}
\mathbf{\tilde{y} = U'y}
\end{equation}

Since the kernel matrix $\mathbf{K}$ has, in general, full rank, the cost of its eigendecomposition is $\mathcal{O}(N^3)$: this is the  initial overhead associated with the proposed set of identities. Propositions \ref{prop1} to \ref{prop3} summarize the main results of the present contribution. Supporting proofs are included in Appendix A.

\begin{proposition} \label{prop1} It holds that

\begin{equation}
\mathcal{L}_y = N\log{\sigma^2} + \sum_{i=1}^N \left( \log{d_i} + \tilde{y}_i^2g_i \right) - 4 \frac{\mathbf{y'y}}{\sigma^2} \label{specscore}
\end{equation}

\noindent where

$$
d_i = \frac{s_i}{s_i + \frac{\sigma^2}{\lambda^2}} + 1 = \frac{2\lambda^2s_i + \sigma^2}{\lambda^2s_i + \sigma^2}
$$

\noindent is the $i$-th eigenvalue of $\mathbf{\Sigma_y}$, and

$$
g_i = \frac{d_i^2 + 4}{\sigma^2d_i} = \frac{8\, {\lambda^4}\, {s_{i}}^2 + 12\, \lambda^2\, s_{i}\, \sigma^2 + 5\, {\sigma^4}}{\sigma^2\, \left(\sigma^2 + \lambda^2\, s_{i}\right)\, \left(\sigma^2 + 2\, \lambda^2\, s_{i}\right)}
$$

\noindent is the $i$-th eigenvalue of $\left(\sigma^{-4}\mathbf{\Sigma_y} + 4\mathbf{\Sigma_y}^{-1}\right)$

 $\qed$

\end{proposition}

Proposition \ref{prop1} exploits the \emph{simultaneously diagonalizable} property of the following pairs of matrices:

\begin{itemize}
\item $\mathbf{K}$ and $\left(\mathbf{K} + \frac{\sigma^2}{\lambda^2}\right)^{-1}$
\item $\mathbf{K}\left(\mathbf{K} + \frac{\sigma^2}{\lambda^2}\right)^{-1}$ and $\left(\mathbf{K}\left(\mathbf{K} + \frac{\sigma^2}{\lambda^2}\right)^{-1} + \mathbf{I}\right)^{-1}$
\end{itemize}

\noindent to compute $\mathcal{L}_y$ as a function of the eigenvalues of $\mathbf{K}$ with a complexity of $\mathcal{O}(N)$.

\begin{proposition} \label{prop2} It holds that

\begin{eqnarray}
\frac{\partial \mathcal{L}_y}{\partial{\sigma^2}} &=& \frac{N}{\sigma^2} + 4\frac{\mathbf{y'y}}{\sigma^4} + \sum_{i=1}^N \left( \frac{\partial \log{d_i}}{\partial\sigma^2} + \tilde{y}_i^2\frac{\partial g_i}{\partial \sigma^2} \right) \label{specscore2}\\
\frac{\partial \mathcal{L}_y}{\partial{\lambda^2}} &=& \sum_{i=1}^N \left( \frac{\partial \log{d_i}}{\partial\lambda^2} + \tilde{y}_i^2\frac{\partial g_i}{\partial \lambda^2} \right)\label{specscore3}
\end{eqnarray}

\noindent with

\begin{eqnarray}
\frac{\partial \log d_i}{\partial \sigma^2} &=& \frac{1}{\sigma^2 + 2\, \lambda^2\, s_{i}} - \frac{1}{\sigma^2 + \lambda^2\, s_{i}}\\
\frac{\partial \log d_i}{\partial \lambda^2} &=& \frac{s_{i}\, \sigma^2}{\left(\sigma^2 + \lambda^2\, s_{i}\right)\, \left(\sigma^2 + 2\, \lambda^2\, s_{i}\right)}
\end{eqnarray}

\noindent and

\begin{eqnarray}
\frac{\partial g_i}{\partial \sigma^2} &=& - \frac{4}{{\sigma^4}} - \frac{{\sigma^8} - 2\, {\lambda^4}\, {s_{i}}^2\, {\sigma^4}}{{\sigma^4}\, {\left(\sigma^2 + \lambda^2\, s_{i}\right)}^2\, {\left(\sigma^2 + 2\, \lambda^2\, s_{i}\right)}^2}\\
\frac{\partial g_i}{\partial \lambda^2} &=& \frac{s_{i}}{{\left(\sigma^2 + \lambda^2\, s_{i}\right)}^2} - \frac{4\, s_{i}}{{\left(\sigma^2 + 2\, \lambda^2\, s_{i}\right)}^2}
\end{eqnarray}

$\qed$

\end{proposition}

\begin{proposition} \label{prop3}

\begin{eqnarray}
\frac{\partial^2 \mathcal{L}_y}{\partial^2 \lambda^2} &=& \sum_{i=1}^N \left( \frac{\partial^2 \log{d_i}}{\partial^2\lambda^2} + \tilde{y}_i^2\frac{\partial^2 g_i}{\partial^2 \lambda^2} \right)\label{specscore4}\\
\frac{\partial^2 \mathcal{L}_y}{\partial \lambda^2\partial \sigma^2} &=& \sum_{i=1}^N \left( \frac{\partial^2 \log{d_i}}{\partial\lambda^2\partial\sigma^2} + \tilde{y}_i^2\frac{\partial^2 g_i}{\partial^2\lambda^2 \partial\sigma^2} \right)\label{specscore5}\\
\frac{\partial^2 \mathcal{L}_y}{\partial^2 \sigma^2} &=& - \frac{N}{\sigma^4} - 8\frac{\mathbf{y}'\mathbf{y}}{\sigma^6} +  \sum_{i=1}^N \left( \frac{\partial^2 \log{d_i}}{\partial^2\sigma^2} + \tilde{y}_i^2\frac{\partial^2 g_i}{\partial^2 \sigma^2} \right)\label{specscore6}\\
\end{eqnarray}

\noindent with

\begin{eqnarray}
\frac{\partial^2 \log d_i}{\partial^2 \lambda^2} &=& \frac{{s_{i}}^2}{{\left(\sigma^2 + \lambda^2\, s_{i}\right)}^2} - \frac{4\, {s_{i}}^2}{{\left(\sigma^2 + 2\, \lambda^2\, s_{i}\right)}^2}\\
\frac{\partial^2 \log d_i}{\partial \sigma^2 \partial \lambda^2} &=& \frac{s_{i}}{{\left(\sigma^2 + \lambda^2\, s_{i}\right)}^2} - \frac{2\, s_{i}}{{\left(\sigma^2 + 2\, \lambda^2\, s_{i}\right)}^2}\\
\frac{\partial^2 \log d_i}{\partial^2 \sigma^2} &=& \frac{1}{{\left(\sigma^2 + \lambda^2\, s_{i}\right)}^2} - \frac{1}{{\left(\sigma^2 + 2\, \lambda^2\, s_{i}\right)}^2}
\end{eqnarray}

\noindent and

\begin{eqnarray}
\frac{\partial^2 g_i}{\partial^2 \lambda^2} &=& \frac{16\, {s_{i}}^2}{{\left(\sigma^2 + 2\, \lambda^2\, s_{i}\right)}^3} - \frac{2\, {s_{i}}^2}{{\left(\sigma^2 + \lambda^2\, s_{i}\right)}^3}\\
\frac{\partial^2 g_i}{\partial\sigma^2\partial\lambda^2} &=& \frac{8\, s_{i}}{{\left(\sigma^2 + 2\, \lambda^2\, s_{i}\right)}^3} - \frac{2\, s_{i}}{{\left(\sigma^2 + \lambda^2\, s_{i}\right)}^3}\\
\frac{\partial^2 g_i}{\partial^2 \sigma^2} &=& \frac{8}{{\sigma^6}} - \frac{12\, {\lambda^6}\, {s_{i}}^3\, {\sigma^6} + 12\, {\lambda^4}\, {s_{i}}^2\, {\sigma^8} - 2\, {\sigma^{12}}}{{\sigma^6}\, {\left(\sigma^2 + \lambda^2\, s_{i}\right)}^3\, {\left(\sigma^2 + 2\, \lambda^2\, s_{i}\right)}^3}
\end{eqnarray}

$\qed$

\end{proposition}

\noindent Propositions \ref{prop2} and \ref{prop3} apply standard calculus rules to compute the Jacobian and Hessian matrices of $\mathcal{L}_y$ with a complexity of $\mathcal{O}(N)$; it should be noted that all the presented quantities are also valid when $\mathbf{K}$ does not have full rank. Versions of Equations (\ref{specscore}), (\ref{specscore2}), (\ref{specscore3}), (\ref{specscore4}), (\ref{specscore5}) and (\ref{specscore6}) that are expressed directly in terms of the eigenvalues $s_i$ are included in Appendix A.  Remarkably, by employing the proposed set of identities, it is possible to compute the current score function, Jacobian vector and Hessian matrix in $\mathcal{O}(N)$ for every iteration of an optimization algorithm, following an initial overhead of $\mathcal{O}(N^3)$ for the eigendecomposition of $\mathbf{K}$. 

The convenient properties of the proposed set of equations allow to assess the uncertainty of the estimated model, described by equation (\ref{Sigma_c}), with reduced computational complexity: this result, whose proof is not reported because of its simplicity, is summarized in the following

\begin{proposition} \label{prop4}

It is apparent that, in order to compute the quantity

\begin{equation}
\mathrm{Var}[\mathbf{c}|\mathbf{y}] = \mathbf{\Sigma_c} = \sigma^2 \left(\mathbf{K} + \frac{\sigma^2}{\lambda^2}\mathbf{I}\right)^{-1}\mathbf{K}^{-1} \label{sigmacinv}
\end{equation}

two $N \times N$ matrices need to be inverted; this results in a cost of $\mathcal{O}(N^3)$. By exploiting again the SVD of $\mathbf{K}$, it follows that

\begin{equation}
\mathbf{\Sigma_c} = \mathbf{U}\mathbf{Q}\mathbf{U}' \label{sigmacmult}
\end{equation}

where $\mathbf{Q}$ is a diagonal matrix whose $i$-th element is

$$
q_i = \frac{\sigma^2\lambda^2}{(\lambda^2s_i + \sigma^2)s_i}
$$

It follows that the matrix $\mathbf{\Sigma_c}$ can be computed using Strassen's algorithm in $\mathcal{O}(N^{\log_2 7}) \approx \mathcal{O}(N^{2.807})$ operations \citep{strassen1969gaussian}. In most cases, only a part of the uncertainty matrix is of interest (for example, only the diagonal elements). It is immediate to notice that, using equation (\ref{sigmacmult}), it is possible to compute directly only the interesting elements (in $\mathcal{O}(N)$ each), where equation (\ref{sigmacinv}) would still require matrix inversion operations of the usual $\mathcal{O}(N^3)$ complexity. $\qed$

\end{proposition}

\subsection{Proof of Computational Advantage}

In order to characterize the advantage that the new formulation brings to the problem at hand, we consider an iterative global optimization technique that is able to solve problem (\ref{scorefun1}) in $k^*$ steps. As discussed in Section \ref{oldoptim}, when the new set of identities is not employed the solution of (\ref{scorefun1}) costs

$$
\mathrm{\tau}_0 = k^*\mathcal{O}(N^3)
$$

\noindent In contrast, when the new set of identities is used, the cost is instead

$$
\mathrm{\tau}_1 = \mathcal{O}(N^3) + k^*\mathcal{O}(N)
$$

\noindent Provided the number of examples, $N$, is such that $N^3 \gg N$, it follows that

\begin{equation}
\frac{\mathrm{\tau}_0}{\mathrm{\tau}_1} = \mathcal{O}(k^*) \label{gain1}
\end{equation}

\noindent Hence, the eigendecomposition formulation results in a speed-up in the solution of (\ref{scorefun1}) by a factor $k^*$. The number of iterations $k^*$ depends, of course, on the characteristics of $\mathcal{S}$, the optimization algorithm employed and the stopping criteria selected. In practice it is common to encounter problems where the value of $k^*$ is in the hundreds. Note that there exists an upper bound on the achievable speed-up:

\begin{equation}
\underset{k^* \rightarrow \infty}{\lim} \frac{\mathrm{\tau}_0}{\mathrm{\tau}_1} = \mathcal{O}(N^2) \label{gain2}
\end{equation}

\noindent For reasonably sized dataset this upper limit has little practical implication (for $N = 200$, equation (\ref{gain2}) would only be relevant if the number of steps needed, $k^*$, is of the order of 40000). By combining (\ref{gain1}) and (\ref{gain2}), the final speed-up order reads

\begin{equation}
\frac{\mathrm{\tau}_0}{\mathrm{\tau}_1} = \mathcal{O}(\min\left\{k^*,\ N^2\right\})
\end{equation}

In order to practically implement the proposed set of equations, memory storage of $\mathcal{O}(N)$ is required: specifically, all the quantities defined in Proposition \ref{prop1}, \ref{prop2} and \ref{prop3} are computable as a function of the eigenvalues $s_i$ and the projected target values $\tilde{y}_i$ (since, following from the properties of Singular Value Decomposition, $\mathbf{\tilde{y}'\tilde{y} = y'y}$). This represents an additional advantage with respect to the procedure described in Section \ref{sec2}, which has $\mathcal{O}(N^2)$ storage requirements. In conclusion, it has been proven that the proposed methodology is able to speed-up hyperparameter tuning operations by many orders of magnitude (up to a factor of $N^2$) and can be implemented with much lower memory requirements than conventional approaches. Furthermore, in the case of multiple-output training datasets, such that

$$
\mathcal{S} = \{\mathbf{X}, \mathbf{y}_1,\ \dots,\ \mathbf{y}_M\}
$$

\noindent the proposed technique has the advantage that the eigendecomposition need only be computed once (since $\mathbf{K}$ depends only on $\mathbf{X}$), to solve the $M$  tuning problems: no additional computational overhead is needed for multiple output-problems. Since state of the art sparse approximations have a computational complexity of $\mathcal{O}(Nm^2)$ per evaluation, where $m^2$ is the number of non-zero elements of the approximation of $\mathbf{K}$ \citep{quiñonero2005unifying}, it is apparent that the proposed set of identities provides a speed-up of hyperparameter tuning even with respect to approximate methods, at least if $k^*$ exceeds a certain threshold that depends on the sparsity rate $m/N$.

\subsection{Kernel Hyperparameters Tuning}

It is quite common for the kernel function $\mathcal{K}$ to depend on additional hyperparameters, such that

$$
\mathcal{K}(x, y) := \mathcal{K}(x, y;\ \theta)
$$

\noindent Notable examples of such parameters are the bandwidth parameter $\xi^2$ of the Radial Basis Function kernel

$$
\mathcal{K}(x, y) = \mathrm{exp}\left(-\frac{||x - y||^2}{2\xi^2}\right)
$$

\noindent and the degree $l$ of the Polynomial kernel

$$
\mathcal{K}(x, y) = (\left\langle x, y \right\rangle + 1)^l
$$

\noindent Since the matrix $\mathbf{K}$ depends on $\theta$, it is in general necessary to recompute the eigendecomposition every time a better value of $\theta$ is found. It is possible, however, to exploit the convenient features of the proposed calculations with an efficient two-step strategy, summarized in the following algorithm.

\mybox{\textbf{Algorithm 1}: Two-step procedure for additional hyperparameter tuning

Let $\theta$ be the additional hyperparameter vector, and let $g_0(\cdot)$ and $g_1(\cdot)$ be implementations of iterative techniques such that

\begin{eqnarray*}
\hat{\theta}_{k+1} &=& g_0(\hat{\theta}_k)\\
\hat{\sigma}^2_{k+1} &=& g_1(\hat{\sigma}^2_{k})\\
\hat{\lambda}^2_{k+1} &=& g_1(\hat{\lambda}^2_{k})
\end{eqnarray*}

where $k$ is the current iteration number. Given suitable initial values $\hat{\theta}_0$, $\hat{\sigma}^2_0$ and $\hat{\lambda}^2_0$, consider the following optimization procedure:

\begin{enumerate}
\item For $k = 0,\dots,k_{max}$
\begin{enumerate}
\item $\hat{\theta}_{k+1} = g_0(\hat{\theta}_k)$
\item For $j = 0, \dots, j_{max}$
\begin{enumerate}
\item $\hat{\sigma}^2_{k+1} = g_1(\hat{\sigma}^2_{k})$
\item $\hat{\lambda}^2_{k+1} = g_1(\hat{\lambda}^2_{k})$
\item Stop if criterion is met
\end{enumerate}
\item End for
\item Stop if criterion is met
\end{enumerate}
\item End for
\end{enumerate}

}

It is then possible to reduce the number of required $\mathcal{O}(N^3)$ operations by implementing an efficient ($\mathcal{O}(N)$) internal loop: this allows a conventional line search to be performed on the "expensive" hyperparameter, while solving the inner loop problems with a much higher efficiency.

\section{SIMULATION RESULTS} \label{sec3}

\begin{figure}[h]
\centering
\includegraphics[width=10cm]{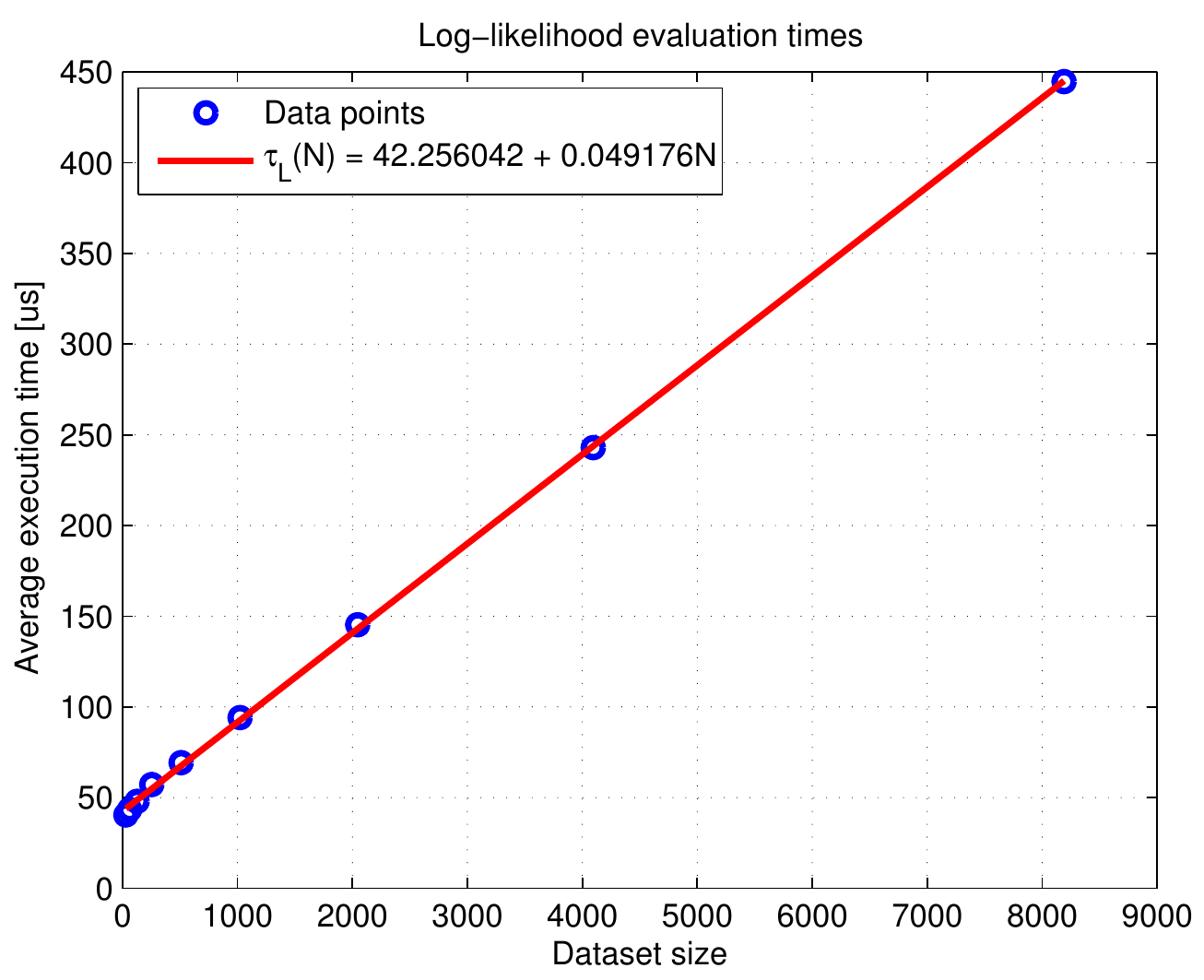}   
\caption{Simulation results for the evaluation of the score function $\mathcal{L}_y$} 
\label{fig:score}
\end{figure}

In order to precisely assess the computational complexity of evaluating the proposed set of identities, a simulation study was conducted for a range of datasets of different sizes. For the sake of reproducibility, we briefly describe the simulation environment: the results were obtained using the MATLAB R2010a numerical engine, running on an Intel(R) Core(TM)2 Quad CPU Q9550 with a clock speed of 2.83GHz, 8 GB of RAM memory and running a 64 bit edition of Windows 7.  Given that Propositions \ref{prop1} to \ref{prop3} predict that $\tau_1(N) = \mathcal{O}(N)$, the goal is to estimate the coefficients of the linear models

$$
\tau(N) = a + bN
$$

\noindent for the computational complexity of the score function evaluation (\ref{specscore}), Jacobian evaluation (\ref{specscore2} and \ref{specscore3}) and Hessian evaluation (\ref{specscore4}, \ref{specscore5} and \ref{specscore6}). In order to obtain a representative dataset, the average execution time (on $10^5$ iterations) of these quantities was evaluated for values of $N$ ranging from 32 to 8192 on a logarithmic scale.

\begin{figure}[h]
\centering
\includegraphics[width=10cm]{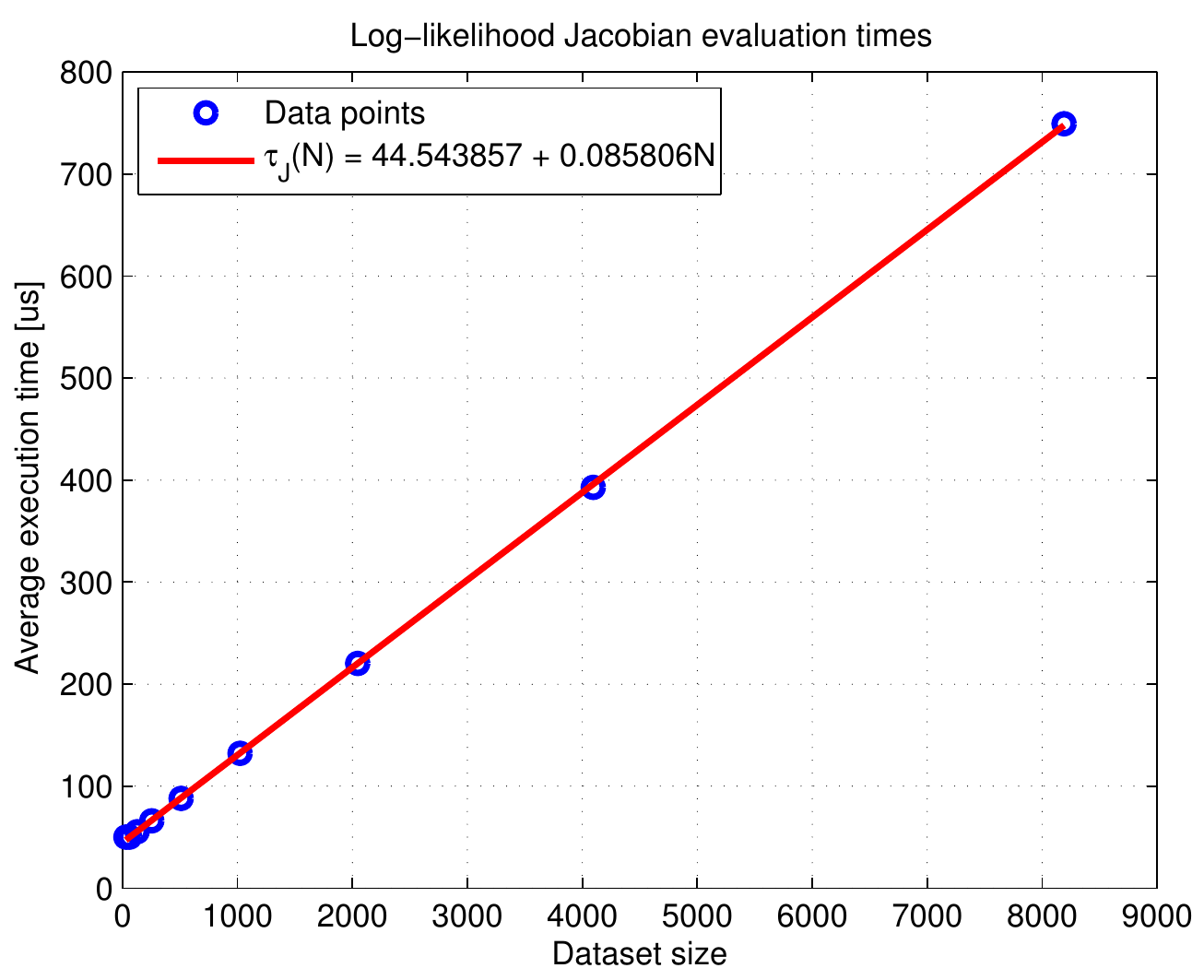}   
\caption{Simulation results for the evaluation of the Jacobian of the score function $\mathcal{L}_y$} 
\label{fig:jac}
\end{figure}

Figure \ref{fig:score} shows the simulation results obtained for the evaluation of the proposed score function equation (\ref{specscore}). Here, the  x-axis is the number of datapoints, $N$, in the dataset and the y-axis is the average computation time (in microseconds). The estimated complexity function is

\begin{equation}
\tau_L(N) \simeq 42.26 + 0.05N\ [\mu s] \label{scorecomp}
\end{equation}

\noindent Remarkably, the computational overhead for evaluating $\mathcal{L}_y$ is only $0.05$ microseconds per observation: This is especially relevant for the global optimization step, which involves a large number of such evaluations.

Figure \ref{fig:jac} shows the simulation results for the estimation of the computational complexity of equations (\ref{specscore2}) and (\ref{specscore3}). The estimated complexity is

\begin{equation}
\tau_J(N) \simeq 44.54 + 0.086N\ [\mu s] \label{jaccomp}
\end{equation}

\noindent As two values need to be computed to build the Jacobian, it is not surprising that the slope coefficient is about twice that of the score function evaluation (equation (\ref{scorecomp})).

\begin{figure}[h]
\centering
\includegraphics[width=10cm]{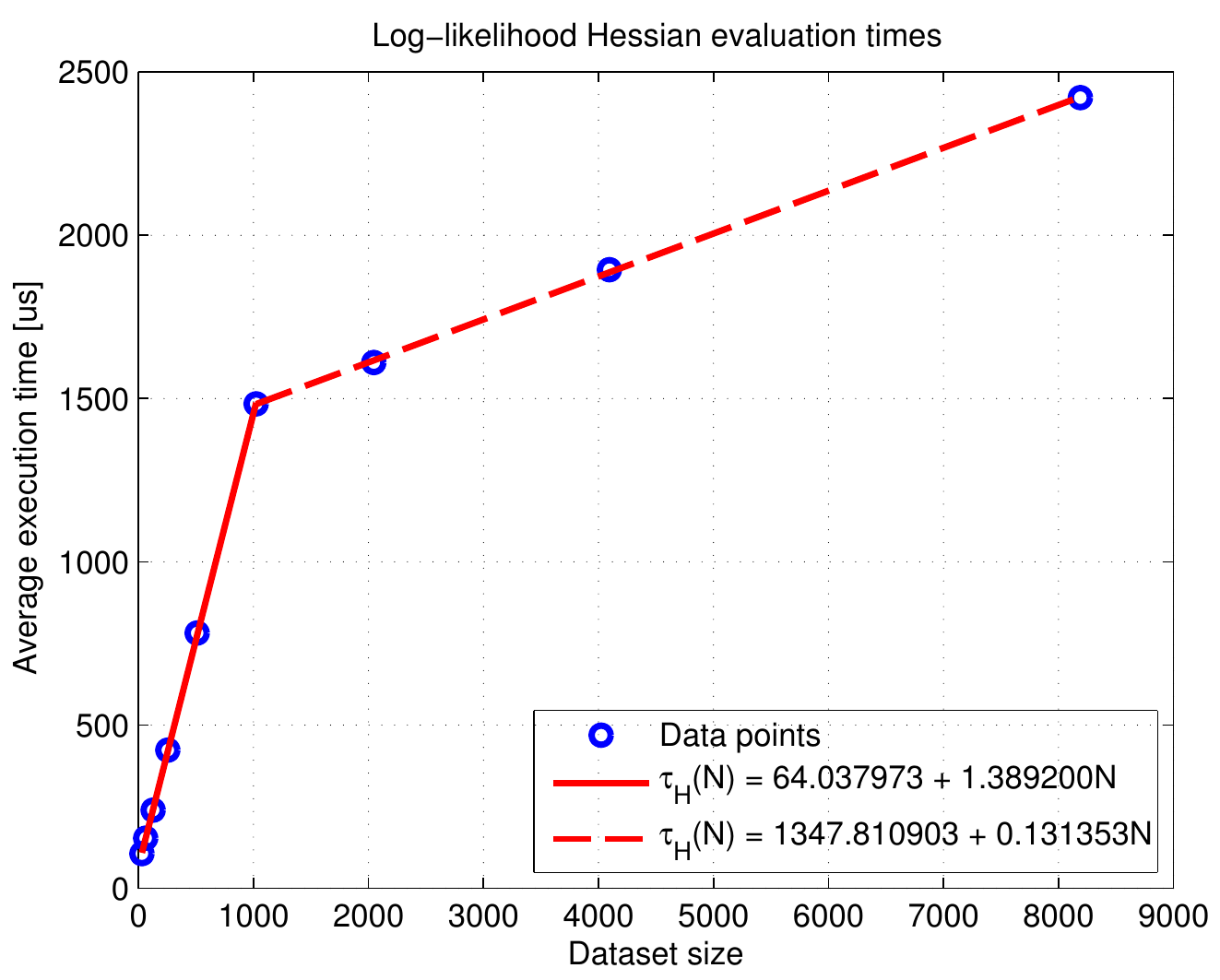}   
\caption{Simulation results for the evaluation of the Hessian of the score function $\mathcal{L}_y$} 
\label{fig:hes}
\end{figure}

The Hessian evaluation, arising from equations (\ref{specscore4}), (\ref{specscore5}) and (\ref{specscore6}), yields somewhat surprising results. As shown in Figure \ref{fig:hes}, the piecewise linear model

\begin{equation}
\tau_{H}(N) \simeq \begin{cases} 64.04 + 1.39N\ [\mu s] & N \leq 1024 \\ 1347.81 + 0.13N\ [\mu s] & N > 1024 \end{cases} \label{hescomp}
\end{equation}

\noindent was needed in order to fit the data. Surprisingly, a large reduction (about an order of magnitude) in the slope is observed for $N > 1024$. As this feature cannot be linked to the theoretical formulation of Proposition \ref{prop3}, the authors conjecture that it relates to internal procedures of the MATLAB numerical engine. Indeed, the simulation experiment was conducted on several computers yielding similar results. It is interesting to observe that the slope of $\tau_H$ for $N > 1024$ is about one and half times the slope of $\tau_J$ and three times the slope of $\tau_L$: this is consistent with the number of unique quantities needed for the Hessian computation.

These simulation results allow the amount of time needed to evaluate all the quantities of the optimization to be estimated. Following on from equations (\ref{scorecomp}), (\ref{jaccomp}) and (\ref{hescomp}) (with $N > 1024$), and assuming a local descent algorithm that makes use of Hessian information, the per iteration computational time for the local optimization step is given by

\begin{equation}
\tau_{LC}(N) \simeq 1434.61 + 0.266N\ [\mu s] \label{LOptComp}
\end{equation}

\noindent while the per iteration computation time for the global optimization step, which depends only on the evaluation of $\mathcal{L}_y$ is given by

\begin{equation}
\tau_{GC}(N) \simeq 44.54 + 0.086N\ [\mu s] \label{GOptComp}
\end{equation}

Thus for a dataset with $N = 8000$ datapoints, for example, the local optimization step computation time is only 3.56 milliseconds per iteration while that of the global optimization step is a mere 440 microseconds. These numbers are even more impressive when one realizes that without the new set of identities, an optimization problem of this size would normally be considered intractable \citep{rasmussen2004gaussianb}.

\section*{CONCLUSIONS}
In practical statistical learning applications, hyperparameter tuning plays a key role in achieving the desired prediction capabilities. Such tuning is usually performed by means of maximization of marginalized posterior likelihood, which presents two main challenges: \textbf{(i)} being a nonconvex optimization problem with local optimal points, global optimization techniques (that are usually demanding in terms of number of function evaluation) must be used, and \textbf{(ii)} such evaluations are usually computationally intensive and scale badly with the number of examples in the training dataset. In the cases considered in this paper, namely kernel ridge regression and Gaussian process regression techniques, evaluating the score function has a complexity of $\mathcal{O}(N^3)$: as a consequence, it is often impractical to tune the hyperparameters using a dataset of sufficient dimensions.

In this paper, a set of new identities derived from a spectral decomposition of the kernel matrix are employed to reduce drastically such complexity. Specifically, after the initial decomposition (that costs $\mathcal{O}(N^3)$), all the quantities involved in the problem are computable in $\mathcal{O}(N)$. This represents an advantage of several orders of magnitude with respect to the state of the art for exact solutions: specifically, it allows the hyperparameter tuning problem to be solved with a speed-up factor $\mathcal{O}(k^*)$, where $k^*$ is the number of required iterations. Furthermore, the required memory storage for the new equations is $\mathcal{O}(N)$, as opposed to $\mathcal{O}(N^2)$. It is then possible to use much larger datasets for hyperparameter tuning purposes, without the need to resort to sparse approximations. A two-stage procedure has also been proposed and discussed, which enables the efficiencies offered by the new identities to be exploited when additional hyperparameters have to be optimized.  Finally simulation results are presented that verify the computational advantages of the new formulation.

\section*{APPENDIX A} \label{Appendix}

Let us consider the following

\begin{theorem} \label{theorem2} Let $\mathbf{A} \in \mathbb{R}^{N \times N}$ and $\mathbf{B}  \in \mathbb{R}^{N \times N}$. If $\mathbf{A}$ and $\mathbf{B}$ commute, such that

$$
\mathbf{A}\mathbf{B} = \mathbf{B}\mathbf{A}
$$

\noindent $\mathbf{A}$ and $\mathbf{B}$ are simultaneously diagonalizable. This means that, given an eigendecomposition of $\mathbf{A}$ such that

$$
\mathbf{U}\mathbf{D_A}\mathbf{U'} = \mathbf{A}
$$

\noindent it holds that

$$
\mathbf{U}\mathbf{D_B}\mathbf{U'} = \mathbf{B}
$$

\noindent where $\mathbf{D_A}$ and $\mathbf{D_B}$ are the diagonal eigenvalues matrices associated to $\mathbf{A}$ and $\mathbf{B}$. $\qed$

\end{theorem}

\begin{proof} of Proposition \ref{prop1}. Considering the identity

$$
(\mathbf{\mu_y} - \mathbf{y}) = \frac{1}{\sigma^2}(\mathbf{\Sigma_y} - 2\sigma^2\mathbf{I})\mathbf{y}
$$

\noindent it follows that, up to an additive constant,

\begin{equation}
\mathcal{L}_y = \log{|\mathbf{\Sigma_y}|} + \frac{1}{\sigma^4}\mathbf{y}'\mathbf{\Sigma_y}\mathbf{y} + 4\mathbf{y}'\mathbf{\Sigma_y}^{-1}\mathbf{y} - 4\frac{\mathbf{y}'\mathbf{y}}{\sigma^2} \label{logl0}
\end{equation}

\noindent In order to efficiently compute $\mathcal{L}_y$ and its derivatives with respect to $\sigma^2$ and $\lambda^2$, consider the eigendecomposition of $\mathbf{K}$ such that

$$
\mathbf{K} = \mathbf{U}\mathbf{S}\mathbf{U}'
$$

\noindent where $\mathbf{U}$ is an orthogonal matrix such that $\mathbf{U}\mathbf{U}' = \mathbf{I}$ and $\mathbf{S}$ is a diagonal matrix whose $i$-th entry $s_i$ is the $i$-th ordered eigenvalue of $\mathbf{K}$. It is easy to prove that $\mathbf{K}$ and $(\mathbf{K} + \frac{\sigma^2}{\lambda^2}\mathbf{I})^{-1}$ commute, and hence

$$
\mathbf{U}\left(\mathbf{S}\left(\mathbf{S} + \frac{\sigma^2}{\lambda^2}\mathbf{I}\right)^{-1}\right)\mathbf{U}' = \mathbf{K}(\mathbf{K} + \frac{\sigma^2}{\lambda^2}\mathbf{I})^{-1}
$$

\noindent Following from (\ref{SigmaY}),

$$
\sigma^2\mathbf{U}\mathbf{D}\mathbf{U}' =  \mathbf{\Sigma_y}
$$

\noindent where the $i$-th entry of the diagonal matrix $D$ is

$$
d_i = \frac{s_i}{s_i + \frac{\sigma^2}{\lambda^2}} + 1 = \frac{2\lambda^2s_i + \sigma^2}{\lambda^2s_i + \sigma^2}
$$

\noindent Recalling that

$$
|\mathbf{A}| = \prod_i \xi_i
$$

\noindent where $\mathbf{A}$ is a square matrix and $\xi_i$ is its $i$-th eigenvalue, it follows that

$$
\log|\mathbf{\Sigma_y}| = \sum_{i=1}^N \log (\sigma^2d_i) = N\log\sigma^2 + \sum_{i=1}^N\log{d_i}
$$

\noindent After this substitution, equation (\ref{logl0}) reads

$$
\mathcal{L}_y = N\log{\sigma^2} + \sum_{i=1}^N \log{d_i} + \frac{1}{\sigma^4}\mathbf{y}'\mathbf{\Sigma_y}\mathbf{y} + 4\mathbf{y}'\mathbf{\Sigma_y}^{-1}\mathbf{y} - 4\frac{\mathbf{y}'\mathbf{y}}{\sigma^2}
$$

\noindent Since $\mathbf{\Sigma_y}$ and $\mathbf{\Sigma_y^{-1}}$ commute, they are simultaneously diagonalizable. Letting

$$
\mathbf{\tilde{y}} = \mathbf{U}'\mathbf{y}
$$

\noindent and letting $\tilde{y}_j$ be the $j$-th element of $\mathbf{\tilde{y}}$,

$$
\mathcal{L}_y = N\log{\sigma^2} + \sum_{i=1}^N \log{d_i} + \mathbf{y}'\mathbf{U}\mathbf{G}\mathbf{U}'\mathbf{y} - 4 \frac{\mathbf{y}'\mathbf{y}}{\sigma^2}
$$

\noindent where $\mathbf{G}$ is a diagonal matrix whose $i$-th entry is

$$
g_i = \frac{d_i^2 + 4}{\sigma^2d_i} = \frac{8\, {\lambda^4}\, {s_{i}}^2 + 12\, \lambda^2\, s_{i}\, \sigma^2 + 5\, {\sigma^4}}{\sigma^2\, \left(\sigma^2 + \lambda^2\, s_{i}\right)\, \left(\sigma^2 + 2\, \lambda^2\, s_{i}\right)}
$$

\noindent The final form of $\mathcal{L}_y$ is then

\begin{eqnarray}
\mathcal{L}_y &=& N\log{\sigma^2} + \sum_{i=1}^N \left( \log{d_i} + \tilde{y}_i^2g_i \right) - 4 \frac{\mathbf{y}'\mathbf{y}}{\sigma^2} \label{finalL}\\
&=& N\log{\sigma^2} - 4 \frac{\mathbf{y}'\mathbf{y}}{\sigma^2} \nonumber\\
&& + \sum_{i=1}^N  \log{\left(\frac{2\lambda^2s_i + \sigma^2}{\lambda^2s_i + \sigma^2}\right)} \nonumber\\
&& + \sum_{i=1}^N  \tilde{y}_i^2\left(\frac{8\, {\lambda^4}\, {s_{i}}^2 + 12\, \lambda^2\, s_{i}\, \sigma^2 + 5\, {\sigma^4}}{\sigma^2\, \left(\sigma^2 + \lambda^2\, s_{i}\right)\, \left(\sigma^2 + 2\, \lambda^2\, s_{i}\right)}\right)\nonumber
\end{eqnarray}

$\qed$

\end{proof}

\begin{proof} of Proposition \ref{prop2}. By applying derivative rules to equation (\ref{finalL}),

\begin{eqnarray*}
\frac{\partial \mathcal{L}_y}{\partial{\sigma^2}} &=& \frac{N}{\sigma^2} + 4\frac{\mathbf{y}'\mathbf{y}}{\sigma^4} - \sum_{i=1}^N  \left( \frac{4\, \tilde{y}_i^2}{{\sigma^4}} + \frac{{\lambda^4}\, \left(3\, {s_{i}}^2\, {\sigma^6} - 2\, {s_{i}}^2\, {\sigma^4}\, \tilde{y}_i^2\right) + {\sigma^8}\, \tilde{y}_i^2 + \lambda^2\, s_{i}\, {\sigma^8} + 2\, {\lambda^6}\, {s_{i}}^3\, {\sigma^4}}{{\sigma^4}\, {\left(\sigma^2 + \lambda^2\, s_{i}\right)}^2\, {\left(\sigma^2 + 2\, \lambda^2\, s_{i}\right)}^2}\right) \\
\frac{\partial \mathcal{L}_y}{\partial{\lambda^2}} &=& -\sum_{i=1}^N  \left(\frac{ - 2\, \sigma^2\, {\lambda^4}\, {s_{i}}^3 + \left(4\, \sigma^2\, \tilde{y}_i^2 - 3\, {\sigma^4}\right)\, \lambda^2\, {s_{i}}^2 + \left(3\, {\sigma^4}\, \tilde{y}_i^2 - {\sigma^6}\right)\, s_{i}}{{\left(2\, {\lambda^4}\, {s_{i}}^2 + 3\, \lambda^2\, s_{i}\, \sigma^2 + {\sigma^4}\right)}^2}\right) 
\end{eqnarray*}

$\qed$

\end{proof}

\begin{proof} of Proposition \ref{prop3}. By applying derivative rules to equation (\ref{finalL}),

\begin{eqnarray*}
\frac{\partial^2 \mathcal{L}_y}{\partial^2 \lambda^2} &=&\sum_{i=1}^N  \left(\frac{ - 8\, \sigma^2\, {\lambda^6}\, {s_{i}}^5 + \left(24\, \sigma^2\, \tilde{y}_i^2 - 18\, {\sigma^4}\right)\, {\lambda^4}\, {s_{i}}^4 + \left(36\, {\sigma^4}\, \tilde{y}_i^2 - 13\, {\sigma^6}\right)\, \lambda^2\, {s_{i}}^3 + \left(14\, {\sigma^6}\, \tilde{y}_i^2 - 3\, {\sigma^8}\right)\, {s_{i}}^2}{{\left(2\, {\lambda^4}\, {s_{i}}^2 + 3\, \lambda^2\, s_{i}\, \sigma^2 + {\sigma^4}^2\right)}^3}\right)\\
\frac{\partial^2 \mathcal{L}_y}{\partial \lambda^2\partial \sigma^2} &=& \sum_{i=1}^N  \left(\frac{4\, {\lambda^8}\, {s_{i}}^5 + \left(6\, \sigma^2 - 8\, \tilde{y}_i^2\right)\, {\lambda^6}\, {s_{i}}^4 + \left(12\, {\sigma^4}\, \tilde{y}_i^2 - 3\, {\sigma^6}\right)\, \lambda^2\, {s_{i}}^2 + \left(6\, {\sigma^6}\, \tilde{y}_i^2 - {\sigma^8}\right)\, s_{i}}{{\left(2\, {\lambda^4}\, {s_{i}}^2 + 3\, \lambda^2\, s_{i}\, \sigma^2 + {\sigma^4}\right)}^3}\right)\\
\frac{\partial^2 \mathcal{L}_y}{\partial^2 \sigma^2} &=& - \frac{N}{\sigma^4} - 8\frac{\mathbf{y}'\mathbf{y}}{\sigma^6} + \sum_{i=1}^N\eta_i\\
\end{eqnarray*}

\noindent with

$$
\eta_i =   \frac{8\, \tilde{y}_i^2}{{\sigma^6}} + \frac{{s_{i}}^2\, \left(9\, {\lambda^4}\, {\sigma^{10}} - 12\, {\lambda^4}\, {\sigma^8}\, \tilde{y}_i^2\right) + {s_{i}}^3\, \left(13\, {\lambda^6}\, {\sigma^8} - 12\, {\lambda^6}\, {\sigma^6}\, \tilde{y}_i^2\right) + 2\, {\sigma^{12}}\, \tilde{y}_i^2 + 2\, \lambda^2\, s_{i}\, {\sigma^{12}} + 6\, {\lambda^8}\, {s_{i}}^4\, {\sigma^6}}{{\sigma^6}\, {\left(2\, {\lambda^4}\, {s_{i}}^2 + 3\, \lambda^2\, s_{i}\, \sigma^2 + {\sigma^4}\right)}^3}
$$

$\qed$

\end{proof}

\bibliographystyle{abbrvnat}
\bibliography{krrtuning}

\end{document}